\documentclass[a4paper,conference]{IEEEtran}
\ifCLASSINFOpdf
  \usepackage[pdftex]{graphicx}
\else
\fi
\usepackage{amsmath}
\usepackage{setspace}

\usepackage{booktabs}

\usepackage{multirow}

\usepackage{array}
\usepackage{epsfig}
\begin{document}
%
\title{Learning Error-Driven Curriculum for Crowd Counting}

\author{\IEEEauthorblockN{Wenxi Li\IEEEauthorrefmark{1},
Zhuoqun Cao\IEEEauthorrefmark{2},
Qian Wang\IEEEauthorrefmark{2},
Songjian Chen\IEEEauthorrefmark{1} and
Rui Feng\IEEEauthorrefmark{1}\IEEEauthorrefmark{2}}
\IEEEauthorblockA{\IEEEauthorrefmark{1}Academy for Engineering and Technology, Fudan University,
Shanghai, China}
\IEEEauthorblockA{\IEEEauthorrefmark{2}Shanghai Key Lab of Intelligent Information Processing, \\
School of Computer Science, Fudan University, Shanghai, China\\
Email: {\{wxli18, zqcao18, wangqian18, sjchen18, fengrui\}@fudan.edu.cn}}
}


\maketitle

\begin{abstract}
  Density regression has been widely employed in crowd counting. 
  However, the frequency imbalance of pixel values in the density map is still an obstacle to improve the performance.
  In this paper, 
  we propose a novel learning strategy for learning error-driven curriculum, which uses an additional network to supervise the training of the main network.
  A tutoring network called TutorNet is proposed to repetitively indicate the critical errors of the main network.
  TutorNet generates pixel-level weights to formulate the curriculum for the main network during training, 
  so that the main network will assign a higher weight to those hard examples than easy examples.
  Furthermore, we scale the density map by a factor to enlarge the distance among inter-examples, which is well known to improve the performance.
  Extensive experiments on two challenging benchmark datasets show that our method has achieved state-of-the-art performance.

\end{abstract}


%
\IEEEpeerreviewmaketitle

\section{Introduction}
\label{sec:intro}
Crowd counting is a heated research topic that employed in many fields, such as video surveillance.
It aims to predict the average densities over the scene image captured by cameras, 
which enables video surveillance system to perform reasoning about both the count and location of the persons.

Most of current state-of-the-art methods for crowd counting are based on density map estimation~\cite{zhang2016single,li2018csrnet}, which uses a fixed Gaussian kernel to normalize a dotted annotation. 
Lempitsk et al.~\cite{lempitsky2010learning} and Zhang et al.~\cite{zhang2016single} proposed the density map generation methods of the fixed convolution kernel and adaptive convolution kernel, respectively. 
In recent years, scholars have designed many different models based on the two kinds of density map mentioned above, continuously reducing errors on several commonly-used benchmark datasets.

\begin{figure}[t]

  \begin{minipage}[b]{.48\linewidth}
    \centering
  \centerline{\epsfig{figure=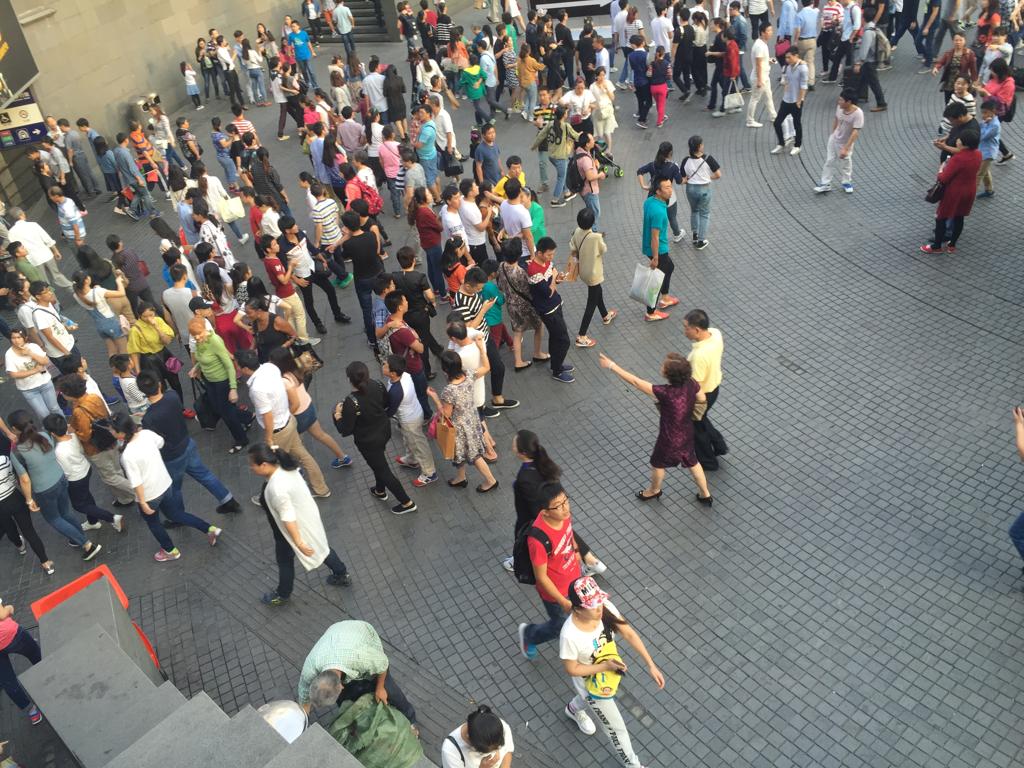,width=3.7cm}}
    \vspace{0cm}
    \centerline{(a) Input image}\medskip
  \end{minipage}
  \hfill
  \begin{minipage}[b]{0.48\linewidth}
    \centering
  \centerline{\epsfig{figure=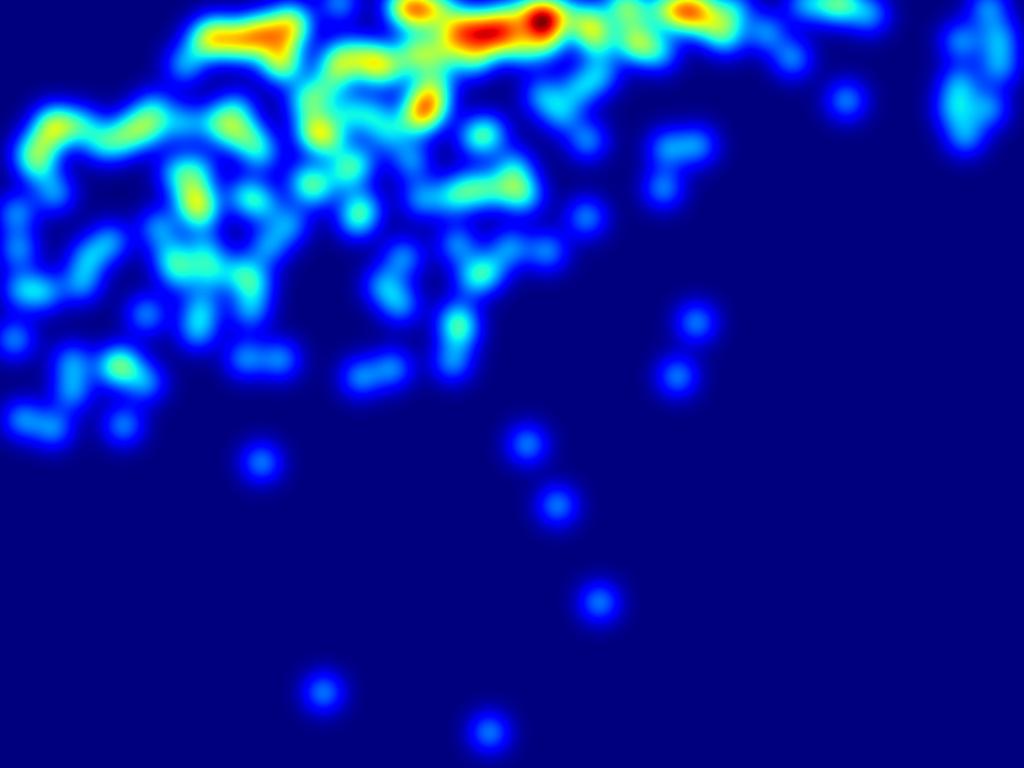,width=3.7cm}}
    \vspace{0cm}
    \centerline{(b) Density map}\medskip
  \end{minipage}
  \begin{minipage}[b]{1.0\linewidth}
    \centering
  \centerline{\epsfig{figure=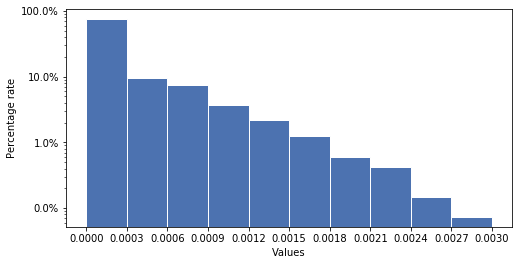,width=8cm}}
    \vspace{0cm}
    \centerline{(c) Hist. of density map}\medskip
  \end{minipage}
  \caption{(a) Input image. (b) Normalize a dotted annotation by a fixed gaussian kernel. (c) The histogram of the
  values in density map.}
  \label{fig:figure1}
  \end{figure}


  Multi-column architectures~\cite{zhang2016single,sam2017switching} and VGG-16~\cite{Simonyan2014Very,li2018csrnet,babu2018divide} are the mainstream backbones for most current methods. 
  A question arises: why the great backbones like ResNet~\cite{he2016deep} or DenseNet~\cite{huang2017densely} seem not so effective for crowd counting?
  We found that the performance of CrowdNet~\cite{boominathan2016crowdnet} and CSRNet~\cite{li2018csrnet} on UCF\_CC\_50~\cite{idrees2013multi} differs greatly, with the mean absolute error of 452.5 and 266.1 respectively, although they all used VGG-16~\cite{Simonyan2014Very} as feature extractor.
  Thus, we can speculate that these mentioned works might be trapped in poor local minima.

  We identify the main bottleneck as the data imbalance issue, which stops density regression from achieving a satisfied accuracy.
  A density map generated through a fixed Gaussian kernel is shown in Figure~\ref{fig:figure1}. 
  From the histogram of the values in the density map, it is quite obvious that there is a serious frequency imbalance of pixel values, and even the peak value in the density map is only 0.003.
  This phenomenon results in very small cost during the training phase, which will hinder the optimization of deep nerual network.
  Furthermore, the distance among different examples is very close, including foreground and background, and different pixel values.
  This also brings challenges to the optimization of the network.

  In this paper, we propose a novel learning strategy to learn from an error-driven curriculum, which can be seen as a particular form of Continuation Method.
  Our learning strategy starts with the target training set and uses a tutoring network called TutorNet to repetitively indicate the errors that the main network makes, which can enhance the effectiveness of learning.
  Specifically, TutorNet generates a weight map to adjust the learning progress of the main network. 
  For a training example with a smaller error, TutorNet will assign a smaller weight to balance the severity of the examples.
  In order to enlarge the distance between different examples,
  we scale the density map by a factor without losing the counting information of the density map.

  
  To evaluate the effectiveness of our learning strategy, four networks previously used for crowd couting are trained under TutorNet's guidance on ShanghaiTech dataset~\cite{zhang2016single}.
  The results show that our approach can significantly improve the performance of different backbone networks.
  Notably, 
  we evaluate our approach with a modified DenseNet~\cite{huang2017densely} as our main network on two challenging benchmark datasets. 
  Compared to current methods, our approach has achieved state-of-the-art performance.

  The main contributions of our work can be summarized as follows: 
  i) We propose a learning strategy for curriculum formulation, which uses weight maps to adjust the learning progress of the main network; 
  ii) We introduce scale factor to enlarge the distance among different examples; 
  iii) Extensive experiments validate the effectiveness of our approach.



\section{RELATED WORK}

\subsection{Deep learning for crowd counting}
Recently, deep learning has greatly stimulated the progress of crowd counting.
Inspired by the great success in deep learning on other tasks, Zhang et al.~\cite{zhang2015cross} focused on the Convolutional Neural Networks (CNN) based approaches to predict the number of crowd firstly.
However, their model ignored the important perspective geometry of scene images and the fully connected layers in them throw away spatial coordinates.
Boominathan et al.~\cite{boominathan2016crowdnet} used the framework of fully convolutional neural networks which combined the deep and shallow networks.
Zhang et al.~\cite{zhang2016single} proposed a multi-column convolutional neural network (MCNN) where each column has a convolution kernel with different sizes for different scales.
Afterward, multi-column architecture became a common way to deal with scale issue~\cite{sam2017switching,sindagi2017generating}. 
For example, Switching-CNN~\cite{sam2017switching} divided each image into non-overlapping patches and used a switch classifier to choose columns for patches.
Since Euclidean loss is sensitive to outliers and have the issue of image blur, ACSCP~\cite{shen2018crowd} used non-overlapping patches and proposed adversarial loss to extend traditional Euclidean loss.
Xiong et al.~\cite{xiong2017spatiotemporal} proposed a model called ConvLSTM and it is the first time incorporating temporal stream for crowd counting.
More recently, Li et al.~\cite{li2018csrnet} proposed a model called CSRNet that used dilated convolution instead of the last two pooling layers of VGG-16~\cite{Simonyan2014Very} and outperformed most of the previous methods.
They developed a single-column network to replace the multi-column network, which has fewer parameters. Liu et al.~\cite{liu2019adcrowdnet} incorporated deformable convolution to address the multi-scale problem.

\subsection{Data imbalance}
The problem of data imbalance, which is common in many vision tasks, has been extensively studied in recent years. 
One way to solve this problem is re-sampling, including the over-sampling method~\cite{chawla2002smote} and down-sampling method~\cite{drummond2003c4}.
Another solution is cost-sensitive learning represented by focal loss~\cite{lin2017focal}, which down-weight the numerous easy negatives.
In fact, the data imbalance problem is rarely studied in regression networks. Lu et al.~\cite{lu2018deep} proposed Shrinkage-loss which takes a step and tries to penalize the easy samples, without decreasing the weight of hard examples.
Although Shrinkage-loss may be helpful in solving data imbalance, the extra hyperparameters need to be manually adjusted, which requires complicated experiment.

\subsection{Continuation methods}
Continuation methods~\cite{allgower1990numerical} tackle optimization problems involving minimizing non-convex criteria.
Curriculum learning~\cite{bengio2009curriculum} is a type of continuation method and self-paced learning~\cite{kumar2010self} is an extension of curriculum learning.
Both of them suggest that samples should be selected in a meaningful order for training.
However, their curriculums need to be predefined. Jiang et al.~\cite{jiang2017mentornet} and Kim et al.~\cite{kim2018screenernet} address this limitation using an attaching network called MentorNet and ScreenerNet respectively.
The difference is that the former uses other datasets for pre-training while the latter uses training data directly.
Inspired by MentorNet and ScreenerNet, we designed an additional network to develop curriculums for the main network of density regression.
Unlike the curriculum learning starting with a small set of easy examples, our learning strategy starts with the target training set and adjust the curriculums according to the severity of the error.

\section{OUR APPROACH}

We propose an effective learning strategy that uses TutorNet to tutor the main network from an error record established during the training phase. 
Moreover, we scale the density maps by a factor to enlarge the distance between different examples during the training phase.

\subsection{TutorNet}
Inspired by the recent success of curriculum learning~\cite{bengio2009curriculum, kumar2010self}, 
we tackle data imbalance by learning a curriculum.
Students build knowledge system in learning, and targeted learning for the loopholes in the knowledge system is more effective than extensive learning. 
In other words, the errors of knowledge point that students make will be repetitively indicated during the curriculum, 
so as to enhance the effectiveness of the learning.  
  There are two important components in our learning strategy.
  One is a main network to generate the density map, the other is a tutoring network.
  We design a TutorNet as the tutoring network to generate the weight map,
  in which each value represents the pixel-level learning rate of the error between the ground truth and density map predicted by the main network.
  The predicted density map and the weight map have the same shape $(W, H)$.
  We define the output weights $w_{x,y}$ bounded between $T$ and 1. 
  When $w_{x,y}$ is close to 1, it means that optimization needs to be continued here.
  Otherwise, when $w_{x,y}$ is close to $T$, it shows that the network has achieved good performance here.
  Formally, we define an activation function to generate $w_{x,y}$:
  \begin{equation}
    \label{equ:activation}
        f\left ( x \right ) = \left\{\begin{matrix}
          \frac{1}{1+e^{-x}}, &if\ x>0
          \\
          \\ T, & if\ x \leq 0 
          \end{matrix}\right., 
    \end{equation}
  where $T$ denotes the adjustable weight for penalizing the easy example. 
  If the weight is too small, it will cause the imbalance to tilt from one extreme to the other. We set $T$ to 0.5 in this paper.

  TutorNet is used to dynamically generate a weight map which is a curriculum for the main network.
  To complete this task, we proposed a loss function to optimize TutorNet:
  \begin{equation}
    \label{eq1}
    \resizebox{.91\linewidth}{!}{$
        \displaystyle
        \mathcal{L}_{Tutor}= \sum_{x}^{H} \sum_{y}^{W} \left ((1-w_{x,y})*e_{x,y}
        +w_{x,y}*max(M-e_{x,y}, 0)  \right ), 
        $}
    \end{equation}
    where $M$ is a margin hyperparameter and $e_{x,y}$ denotes the error computed between the predicted density map and the ground truth.
    We use mean squared error (MSE) for $e_{x,y}$ in this paper.
    The visualization of our loss function is shown in Figure~\ref{fig:lossfig}.
    The gradient of the objective function is given by:
  
    \begin{figure}[t]
      \begin{minipage}[b]{1.0\linewidth}
        \centering
      \centerline{\epsfig{figure=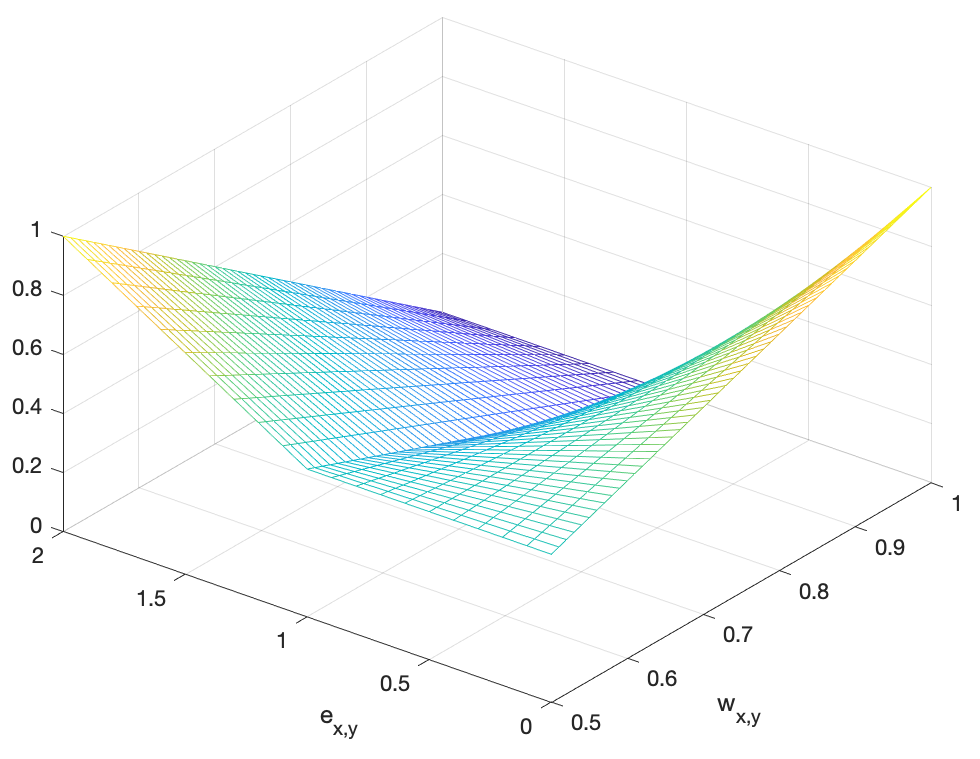,width=8cm}}
        \vspace{0cm}
      \end{minipage}
      \caption{Loss function $\mathcal{L}_{Tutor}$ with M = 1. The error $e_{x,y}$ is given by main network and used to optimize the $w_{x,y}$ from TutorNet.}
      \label{fig:lossfig}
      \end{figure}

    \begin{equation}
      \label{daoshu}
          \frac{\partial \mathcal{L}_{Tutor} }{\partial w_{x,y}} = \left\{\begin{matrix}
            M-2e_{x, y}, & if \ e_{x,y}<M\\ 
            \\ -e_{x,y}, & if \ e_{x,y} \geq M\
            
            \end{matrix}\right., 
      \end{equation}
  
\begin{figure}[t]
      \begin{minipage}[b]{1.0\linewidth}
        \centering
      \centerline{\epsfig{figure=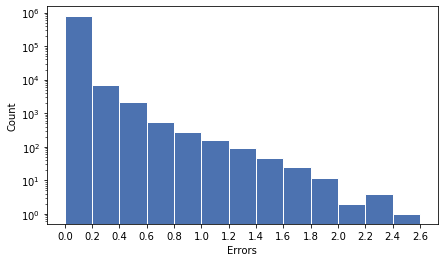,width=8cm}}
        \vspace{0cm}
      \end{minipage}
      \caption{Hist. of the error between the ground truth and the main network prediction.}
      \label{fig:error}
      \end{figure}

    This non-negative error is given by main network and only used to optimize the TutorNet. 
    A gradient descent step is 
    \begin{equation}
      \label{equ:youhua}
      w_{x,y} := w_{x,y} - \alpha  \frac{\partial \mathcal{L}_{Tutor} }{\partial w_{x,y}}  ,
    \end{equation}
    where $\alpha$ is learning rate.
    As the gradient shown above, when the error $e_{x,y}$ is larger than $\frac{M}{2}$, the gradient is a negative value which will increase the value of $w_{x,y}$.
    When the error $e_{x,y}$ is less than $\frac{M}{2}$, the weight $w_{x,y}$ begin its descent.
    We set the $M$ to be 0.8 when using Gaussian kernel (sigma=15) to generate ground truth. 
    The error of one sample during training phase is shown in Figure~\ref{fig:error}.
    
    \begin{figure*}[t]
      
      \begin{minipage}[b]{1.0\linewidth}
        \centering
      \centerline{\epsfig{figure=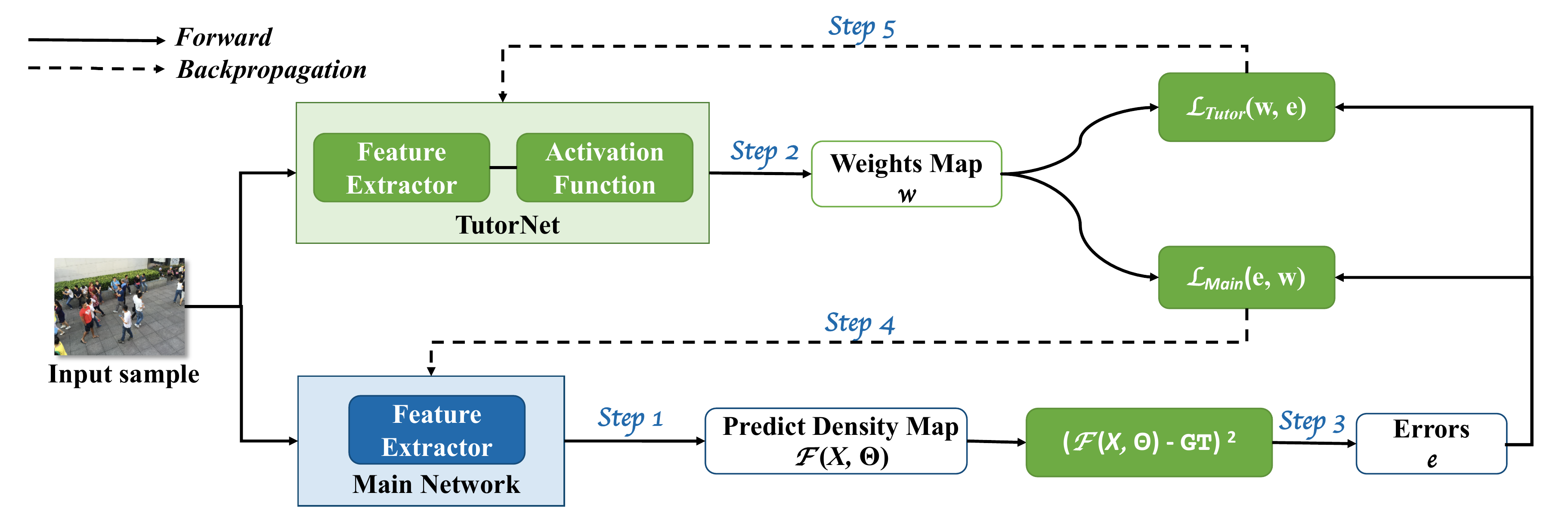,width=15.5cm}}
        \vspace{0cm}
      \end{minipage}
      \caption{Overview of the our learning strategy.}
      \label{fig:network}
      \end{figure*}

    Optimization path of both TutorNet and the main network is shown in Figure~\ref{fig:network}. The loss function to optimize the main network is:
    \begin{equation}
      \label{eq2}
      \resizebox{.91\linewidth}{!}{$
          \displaystyle
          \mathcal{L}_{Main}= { \frac{1}{H*W}\sum_{x}^{H}\sum_{y}^{W} \left ( \mathcal{F}(X,\Theta )_{x,y} - GT_{x,y} \right )^2*w_{x,y} } , 
          $}
      \end{equation}
      where the function $\mathcal{F} (X, \Theta)$ represents the main network and $\Theta$ is the parameters of it. $X$ denotes the input image and $GT$ is the ground truth of it.

  \textbf{TutorNet architecture:} 
  Our TutorNet is used for a tutorial during the training phase.
  The network needs to be able to quickly find the corresponding $w_{x,y}$ that conforms to the current sample. Otherwise, it will reduce the performance of the main network. 
  Our experiment shows that a pre-trained ResNet~\cite{he2016deep} is more likely to converge quickly and achieve good performance.
  Finally, we design TutorNet based on ResNet and we will evaluate TutorNet with different depths in Section~\ref{sec:TutorNet ablation}. 

  \subsection{Density map with scale factor}
  \label{sec:Scale Factor approach}
  
  As we introduced in Section~\ref{sec:intro}, the values in the density map are very small,
  leading to a smaller distance between foreground and background examples which is not conducive to network learning.
  However, it is a common knowledge that excellent training samples are characterized by the large distance among inter-class and the small distance among intra-class.
  Obviously, it is necessary to enlarge the values in the density map.
  
  Note that simply normalizing each density map is not available. 
    The count of objects is given by integral over any image region and simply normalizing will change what the density map can represent.
    We address the problem through scaling density map by a factor, which could linearly transform ground truth by amplifying the values of the non-zero region.
   More experiments about choosing a scale factor will be introduced in Section~\ref{sec:Scale Factor}.

    \begin{figure}[t]
  
      \begin{minipage}[b]{1.0\linewidth}
        \centering
      \centerline{\epsfig{figure=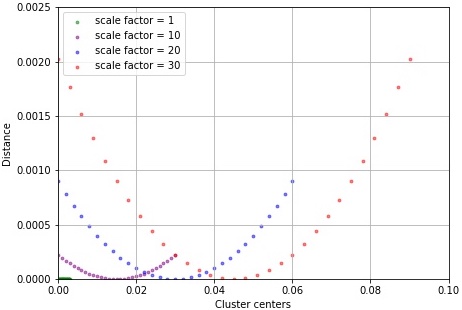,width=7cm}}
        \vspace{0cm}
      \end{minipage}
      \caption{The distance of each group to the mean of different scale factor.}
      \label{fig:scalefactor}
      \end{figure}
  
    From another more intuitive perspective, we approximate each value in density map to four decimal places to achieve clustering.
    We neglect the number of samples in each group, only calculate the mean of these groups. 
    In other words, we calculate the mean of the values represented by each group.
    Then, the euclidean distance of each group to this mean is calculated as shown in Figure~\ref{fig:scalefactor}.
    It can be seen that the scale factor can effectively enlarge the distance, which makes it easier for the network to distinguish different examples. 
    In particular, the scale factor can force the network to fit the foreground (non-zero region) rather than the background (zero region).

\section{EXPERIMENTS}
We first perform ablation experiments to determine the appropriate scale factor and effective TutorNet structure.
Then our best experimental results are compared with the most state-of-the-art methods.

\subsection{Evaluation Metric}
\label{sec:maemse}
The count error is measured using two metrics: Mean Absolute Error (MAE) and Mean Squared Error (MSE).
The MAE is defined as
\begin{equation}
    MAE = \frac{1}{N}\sum_{i=1}^{N}\left | C_{x_{i}} - C_{x_{i}}^{GT} \right |, 
\end{equation}
and the MSE is defined as
\begin{equation}
    MSE =\sqrt[]{ \frac{1}{N}\sum_{i=1}^{N} \left ( C_{x_{i}} - C_{x_{i}}^{GT}  \right )^2}, 
\end{equation}
where $N$ is the number of images in test set, $ C_{x_{i}} $ is the estimated number of people, $ C_{x_{i}}^{GT} $ is the number of 
people in ground truth. 

\subsection{Datasets}
We evaluate our models on ShanghaiTech~\cite{zhang2016single} and Fudan-ShanghaiTech (FDST) video crowd counting dataset~\cite{fang2019locality}, 
because both datasets are captured by surveillance cameras. 

\subsubsection{ShanghaiTech dataset}
We first evaluate our method on the ShanghaiTech~\cite{zhang2016single}. It contains 1,198 annotated images with
a total of 330,165 people with centers of their heads annotated.
The dataset consists of two parts, including Part A and Part B.
The images in Part A are from the internet, and those in Part B are from the busy streets of metropolitan areas in Shanghai.
There are more congested scenes in Part A and more sparse scenes in Part B.
We choose the Part B, which is closer to the real scene, for the experiment.
In Part B, 716 images are used for training and 400 images for testing.

\subsubsection{Fudan-ShanghaiTech dataset}
FDST dataset~\cite{fang2019locality} is the largest video crowd counting dataset.
It contains 100 video sequences captured by 13 surveillance cameras with different scenes.
There are 9,000 annotated frames in the training set, which are from 60 videos. 
The test set consists of 40 videos, 6000 frames.

\subsection{Ablation Study}
\label{sec:Ablation Study}

\subsubsection{Scale Factor}
\label{sec:Scale Factor}
As we introduced in Section~\ref{sec:Scale Factor approach}, the scale factor is used to linearly transform the density map.
For this purpose, we evaluated the performance of different scale factors with DenseNet~\cite{huang2017densely} as shown in Table~\ref{tab:scalefactor}.
The experimental results show the highly performance when the scale factor is 1000. 
Therefore, we set the scale factor to be 1000 for the following experiments.

\begin{table*}[t]
  \caption{The architecture of TutorNet. Downsampling is performed by the first convolution in residual block(2) with a stride of 2. 
  The downsampling rate for all the networks is 8.} 
  \label{tab:tutor architecture}
  \centering
  \begin{tabular}{|c|c|c|c|c|c|}
    \hline
    layer name & 15-layer & 29-layer & 43-layer & 94-layer & layer stride\\ 
    \hline
    convolution & \multicolumn{4}{c|}{7$\times$7, 64, stride 2} & 2\\
    \hline 
    pooling & \multicolumn{4}{c|}{3$\times$3 max pool, stride 2} & 2\\
    \hline
  \multirow{4}{*}{residual block(1)}& \multirow{4}{*}{ $  \begin{bmatrix} 3\times3,64\\ 3\times3,64\\  \end{bmatrix} \times 2 $ } & \multirow{4}{*}{ $  \begin{bmatrix} 3\times3,64\\ 3\times3,64\\  \end{bmatrix} \times 3 $ } & \multirow{4}{*}{ $  \begin{bmatrix} 1\times1,64\\ 3\times3,64\\ 1\times1,256\\ \end{bmatrix} \times 3 $ } & \multirow{4}{*}{ $  \begin{bmatrix} 1\times1,64\\ 3\times3,64\\ 1\times1,256\\ \end{bmatrix} \times 3 $} & \multirow{4}{*}{1}\\
    &&&&&\\
    &&&&&\\
    &&&&&\\
    \hline
  \multirow{4}{*}{residual block(2)}& \multirow{4}{*}{ $  \begin{bmatrix} 3\times3,128\\ 3\times3,128\\  \end{bmatrix} \times 2 $ } & \multirow{4}{*}{ $  \begin{bmatrix} 3\times3,128\\ 3\times3,128\\  \end{bmatrix} \times 4 $ } & \multirow{4}{*}{ $  \begin{bmatrix} 1\times1,128\\ 3\times3,128\\ 1\times1,512\\ \end{bmatrix} \times 4 $ } & \multirow{4}{*}{ $  \begin{bmatrix} 1\times1,128\\ 3\times3,64\\ 1\times1,512\\ \end{bmatrix} \times 4 $} & \multirow{4}{*}{2}\\
    &&&&&\\
    &&&&&\\
    &&&&&\\ 
    \hline
    \multirow{4}{*}{residual block(3)} & \multirow{4}{*}{ $  \begin{bmatrix} 3\times3,256\\ 3\times3,256\\  \end{bmatrix} \times 2 $ } & \multirow{4}{*}{ $  \begin{bmatrix} 3\times3,256\\ 3\times3,256\\  \end{bmatrix} \times 6 $ } & \multirow{4}{*}{ $  \begin{bmatrix} 1\times1,256\\ 3\times3,256\\ 1\times1,1024\\ \end{bmatrix} \times 6 $ } & \multirow{4}{*}{ $  \begin{bmatrix} 1\times1,256\\ 3\times3,256\\ 1\times1,1024\\ \end{bmatrix} \times 23 $} & \multirow{4}{*}{1}\\
      &&&&&\\
    &&&&&\\
    &&&&&\\ 

      \hline  
  \multirow{2}{*}{convolution} &  \multirow{2}{*}{1$\times$1, 1} & \multirow{2}{*}{1$\times$1, 1} & 1$\times$1, 128 &1$\times$1, 128 & \multirow{2}{*}{1}\\
  \cline{4-5}
   &&& 1$\times$1, 1&1$\times$1, 1 & \\
    
    \hline
  \end{tabular}
  \end{table*}

\begin{table}[t]
  \begin{center}
  \caption{Comparison of different scale factors on the ShanghaiTech Part B dataset.} \label{tab:scalefactor}
  \begin{tabular}{|p{3.2cm}|p{2cm}<{\centering}|p{2cm}<{\centering}|}
    \hline
    Scale Factor & MAE & MSE
    \\
    \hline
    \hline
    Density Map * 1 &13.0 & 22.7 \\
    Density Map * 10 &8.3 & 15.0 \\
    Density Map * 100 & 8.2 & 15.8\\
    Density Map * 1000 & \textbf{7.5} & \textbf{12.8} \\
    Density Map * 2000 & 7.6 & 13.7 \\
    \hline
  \end{tabular}
  \end{center}
  \end{table}

  \begin{table}[t]
    \begin{center}
    \caption{Comparison of different TutorNet with scale factor 1000 on the ShanghaiTech Part B dataset.} 
    \label{tab:tutornet}
    \begin{tabular}{|p{3.2cm}|p{2cm}<{\centering}|p{2cm}<{\centering}|}
      \hline
      Method & MAE & MSE
      \\
      \hline
      \hline
      Baseline &7.5 & 12.8 \\
      \hline
      Baseline+15-layer &7.6 & 13.0 \\
      Baseline+29-layer &7.3 & 13.1 \\
      Baseline+43-layer & \textbf{7.0} & \textbf{12.2} \\
      Baseline+94-layer & 7.4 & 12.8 \\
      \hline
    \end{tabular}
    \end{center}
    \end{table}

\begin{table}[t]
  \begin{center}
  \caption{Comparison among various methods with our learning strategy on the ShanghaiTech B dataset. SF and TN respectively refer to the scale factor and TutorNet.} \label{tab:ablation}
  \begin{tabular}{|p{3.2cm}|p{2cm}<{\centering}|p{2cm}<{\centering}|}
    \hline
    Method & MAE & MSE
    \\
    \hline
    \hline
    MCNN~\cite{zhang2016single} & 26.4 & 41.3 \\
    MCNN+SF & 15.3 & 35.2 \\
    MCNN+SF+TN & 14.4 & 25.1 \\
    \hline
    CSRNet~\cite{li2018csrnet} & 10.6 & 16.0 \\
    CSRNet+SF & 10.4 & 15.9 \\
    CSRNet+SF+TN & 9.4 & 15.6 \\
    \hline
    U-net~\cite{ronneberger2015u} & 26.8 & 39.7 \\
    U-net+SF& 13.5 & 23.0 \\
    U-net+SF+TN & 12.1 & 19.7 \\
    \hline
    DenseNet~\cite{huang2017densely} & 13.0 & 22.7 \\
    DenseNet+SF & 7.5 & 12.8 \\
    DenseNet+SF+TN & \textbf{7.0} & \textbf{12.2} \\
    \hline
  \end{tabular}
  \end{center}
  \end{table}

  \subsubsection{TutorNet}
  \label{sec:TutorNet ablation}
  The configurations of TutorNet we designed are shown in Table~\ref{tab:tutor architecture}. 
  We perform experiments on several TutorNet architecture using 1000 as the scale factor.
  The baseline is the main network only~(DenseNet with scale factor 1000) which is the same as Section~\ref{sec:Scale Factor}.
  As we know, an excellent tutor knows how to adjust the learning progress, neither neglecting the mastery of students nor having a lot of repeated learning. 
  The same is true of our TutorNet. 
  It is difficult for a shallow network to master the learning progress of the main network, and the slow convergence speed caused by a very deep network is also an obstacle to the main network.
  The experimental results which are shown in Table~\ref{tab:tutornet} confirm our analysis.
  Therefore, we use the TutorNet with 43 layers for the following experiments.

  \subsubsection{Different architecture}
\begin{figure*}[t]
      
      \begin{minipage}[b]{1.0\linewidth}
        \centering
      \centerline{\epsfig{figure=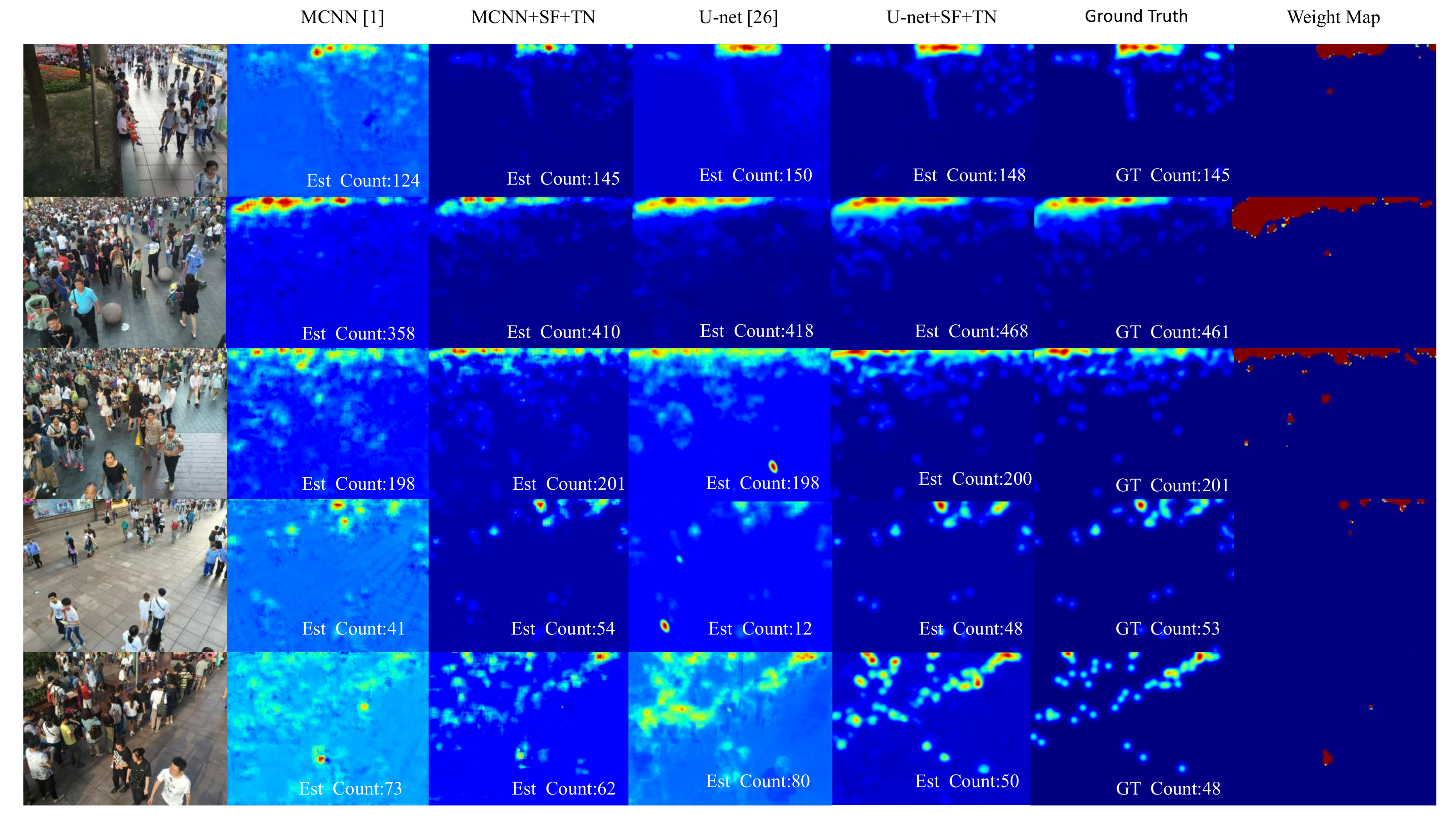,width=17.5cm}}
        \vspace{0cm}
      \end{minipage}
      \caption{Samples generated by different model from ShanghaiTech Part B~\cite{zhang2016single}.}
      \label{fig:visual}
      \end{figure*}

  The ablation study are shown in Table~\ref{tab:ablation}. 
  We choose four typical network architectures as main network to demonstrate our method, 
  a multi-column network called MCNN~\cite{zhang2016single}, 
  a VGG-based network with pre-training called CSRNet~\cite{li2018csrnet},
  a fully convolutional network called U-Net~\cite{ronneberger2015u}
  and a very deep convolutional network DenseNet~\cite{huang2017densely}.
  Scale Factor and TutorNet are individually added to the model training process.
  The experiments verify the effectiveness of our methods in overcoming the data imbalance.
  To evaluate the quality of generated density map, we compare original methods and those methods trained with scale factor and TutorNet. 
  Samples of the test cases can be found in Figure~\ref{fig:visual}.
  The weight maps in the figure are generated during the training phase of MCNN~\cite{zhang2016single} .
  We can see that our method can generate more accurate density maps than the individual main network.

\subsection{Comparison with the state-of-the-art method}
\subsubsection{Experiments on ShanghaiTech}
Table~\ref{tab:soft} compares the performance of our best approach, DenseNet+SF+TN, with state-of-the-art methods.
The results indicate that a DenseNet-based network outperforms most of the previous methods.

  \begin{table}[t]
    \begin{center}
    \caption{Comparison of our best approach with state-of-the-art on the ShanghaiTech Part B dataset.} \label{tab:soft}
    \begin{tabular}{|p{3cm}|p{2cm}<{\centering}|p{2cm}<{\centering}|}
      \hline
      Method & MAE & MSE
      \\
      \hline
      \hline
      MCNN~\cite{zhang2016single} & 26.4 & 41.3 \\
      \hline
      Switching-CNN~\cite{sam2017switching}   & 21.6  & 33.4\\
      \hline
      L2R~\cite{liu2018leveraging}  & 13.7  & 21.4\\
      \hline
      ACSCP~\cite{shen2018crowd}  & 17.2  & 27.4\\
      \hline
      DRSAN~\cite{liu2018crowd}   & 11.1  & 18.2\\
      \hline
      IG-CNN~\cite{babu2018divide}    & 10.7  & 16.0\\ 
      \hline
      CSRNet~\cite{li2018csrnet} & 10.6 & 16.0 \\
      \hline
      ADCrowdNet~\cite{liu2019adcrowdnet} & 7.6 & 13.9 \\
      \hline
      BL~\cite{ma2019bayesian} & 7.7 & 12.7 \\
      \hline
      Ours & \textbf{7.0} & \textbf{12.2} \\
      \hline
    \end{tabular}
    \end{center}
    \end{table}

    \subsubsection{Experiments on FDST dataset}
    Our best approach, DenseNet+SF+TN, is evaluated against a method for single image crowd counting called MCNN~\cite{zhang2016single} and three methods for video crowd counting called ConvLSTM~\cite{xiong2017spatiotemporal},
    LSTN w/o LST~\cite{fang2019locality}  and LSTN~\cite{fang2019locality} respectively.
    The latter three methods exploit the spatial-temporal consistency between frames.
    The results are shown in Table \ref{tab:soft_fudan}.
    Our model outperform the previous methods with only exploiting the spatial information.

  \begin{table}[t]
    \begin{center}
    \caption{Comparison of our best approach with state-of-the-art on the FDST dataset.} \label{tab:soft_fudan}
    \begin{tabular}{|p{3cm}|p{2cm}<{\centering}|p{2cm}<{\centering}|}
      \hline
      Method & MAE & MSE
      \\
      \hline
      \hline
      MCNN~\cite{zhang2016single} & 3.77 & 4.88 \\
      \hline
      ConvLSTM~\cite{xiong2017spatiotemporal}   & 4.48  & 5.82\\
      \hline
      LSTN w/o LST~\cite{fang2019locality}   & 3.87  & 5.16\\
      \hline
      LSTN~\cite{fang2019locality}   & 3.35  & 4.45\\
      \hline
      Ours & \textbf{3.05} & \textbf{4.30} \\
      \hline
    \end{tabular}
    \end{center}
    \end{table}

\section{Conclusion}
  In this work, we identify that data imbalance is the primary obstacle impeding the crowd counting method from achieving state-of-the-art accuracy. 
  Thus, we propose the TutorNet for curriculum formulation, which uses weight maps to adjust the learning progress of the main network.
  Experiments in Section~\ref{sec:Scale Factor} validate the effect of scale factor and our analysis in Section~\ref{sec:Scale Factor approach}.
  An ablation study on four architectures verify the effectiveness of our methods on data imbalance.
  It is worth mentioning that our approach can be easily extended to any other previous network and future network.
  Future work will focus on exploiting the spatial-temporal consistency between video frames.

\section*{Acknowledgment}
This work was supported by Military Key Research Foundation Project (No. AWS15J005), National Natural Science Foundation of China (No. 61672165 and No. 61732004), Shanghai Municipal Science and Technology Major Project (2018SHZDZX01) and ZJLab.

\newpage
\IEEEtriggeratref{15}

\bibliographystyle{IEEEtran}
\bibliography{icme2020template}
%



\end{document}